\documentclass[]{arxiv_article}

\usepackage[toc,page,header]{appendix}
\microtypesetup{expansion=false}
\usepackage{xspace}
\usepackage{enumitem}
\usepackage[linesnumbered,ruled,vlined]{algorithm2e}
\usepackage{amsmath}
\usepackage{amssymb}
\definecolor{darkgreen}{rgb}{0.0,0.45,0.0}

\newcommand{\keywords}[1]{\par\medskip\noindent\textbf{Keywords:}\space\def\and{, }#1}

\title{dWorldEval: Scalable Robotic Policy Evaluation via \\ Discrete Diffusion World Model}
\author[1]{Yaxuan Li}{}
\author[1]{Zhongyi Zhou}{}
\author[1]{Yefei Chen}{}
\author[1]{Yaokai Xue}{}
\author[2]{Yichen Zhu}{}

\affiliation[1]{Current Robotics}
\affiliation[2]{University of Toronto}

\abstract{
Evaluating robotics policies across thousands of environments and thousands of tasks is infeasible with existing approaches. This motivates the need for a new methodology for scalable robotics policy evaluation. In this paper, we propose dWorldEval, which uses a discrete diffusion world model as a scalable evaluation proxy for robotics policies. Specifically, dWorldEval maps all modalities—including vision, language, and robotic actions—into a unified token space, modeling them via a single transformer-based denoising network. 
In this paper, we propose dWorldEval, using a discrete diffusion world model as a scalable evaluation proxy for robotics policy. Specifically, it maps all modalities, including vision, language, and robotics action into a unified token space, then denoises them with a single transformer network. Building on this architecture, We employ a sparse keyframe memory to maintain spatiotemporal consistency. We also introduce a progress token that indicates the degree of task completion. At inference, the model jointly predicts future observations and progress token, allowing automatically determine success when the progress reaches 1. Extensive experiments demonstrate that dWorldEval significantly outperforms previous approaches, i.e., WorldEval, Ctrl-World, and WorldGym, on LIBERO, RoboTwin, and multiple real-robot tasks. It paves the way for a new architectural paradigm in building world simulators for robotics evaluation at scale.

}

\date{\today}
\checkdata[Project Page]{\url{https://dworldeval.github.io/}}

\teaser{
    \begin{center}
        \includegraphics[width=0.9\textwidth]{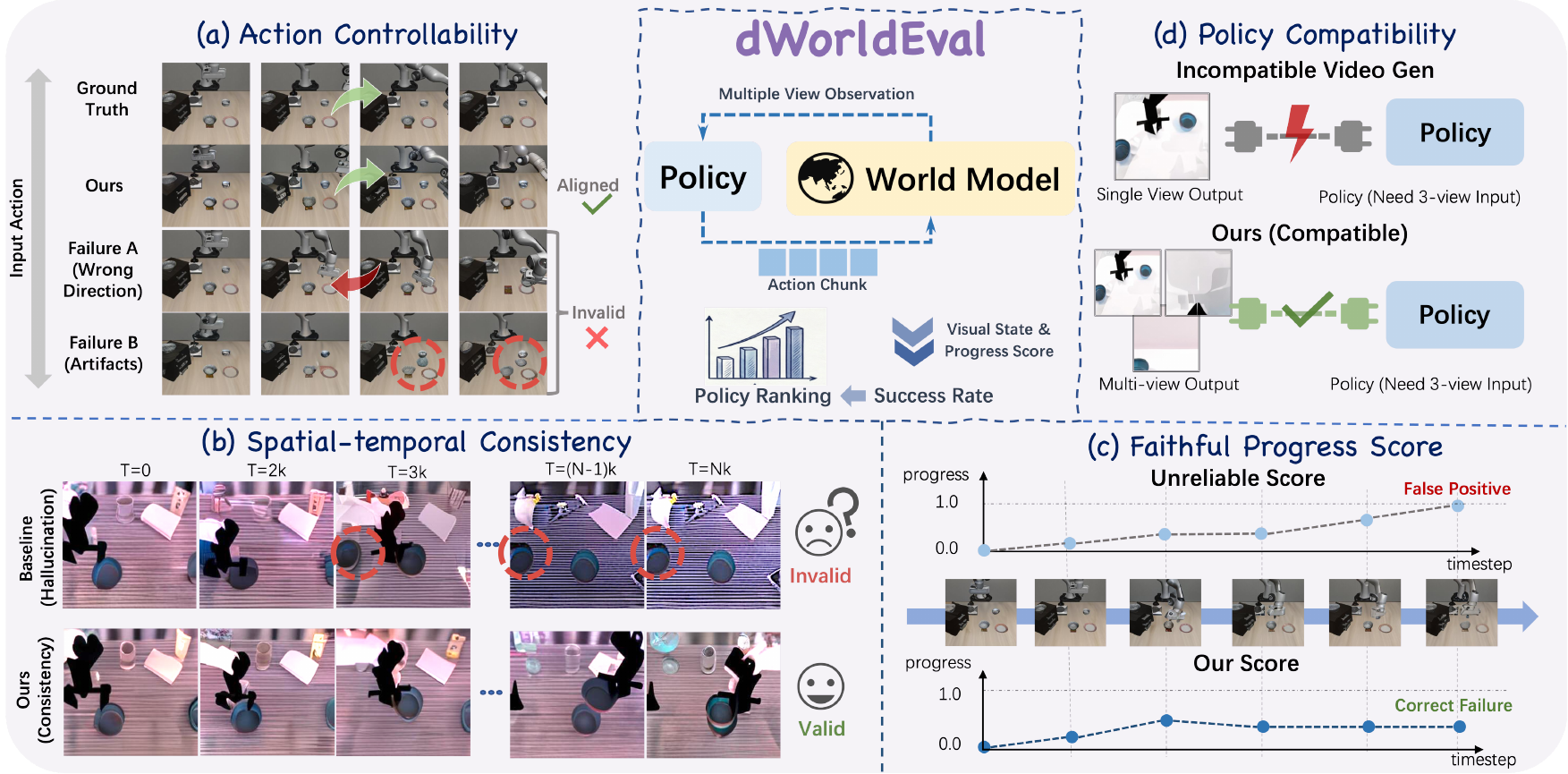} 
        \captionof{figure}{dWorldEval is a discrete diffusion world model designed to be controllable and consistent for reliable policy evaluation. 
    It ensures action controllability and spatiotemporal consistency while providing the policy compatibility and faithful progress scoring necessary for accurate, automated policy ranking.} 
        \label{fig:firstfig}
    \end{center}
}

\begin{document}
\begingroup
\hypersetup{linkcolor=black}
\maketitle
\endgroup


\section{Introduction}
While generalist robot manipulation policies have advanced rapidly~\citep{[pi0, rt-2, kim24openvla, hu2023look, liu2025hybridvla, intelligence2025pi_, kim2025openvlaoft, team2025gemini, bjorck2025gr00t, zhao2025cot-vla, zhao2025vlas, zhen20243d}, evaluating their capabilities remains a significant challenge. 
Consequently, generative world models have emerged as a scalable alternative to costly real-world execution or asset-heavy simulations for evaluating robot policies
~\cite{li2025worldeval,guo2025ctrl,quevedo2506worldgym,team2025evaluating,ho20251x,huang2025enerverse,jiang2025enerverse,tseng2025scalable}.

However, world models have not yet become reliable evaluation proxies for robotics policies, primarily because they often fail to accurately reflect robot actions and physical interactions. We attribute these failures to two main causes. First, existing models struggle to generalize to out-of-distribution (OOD) actions. Since they are typically trained on successful demonstrations, they tend to ignore erroneous actions and hallucinate successful outcomes due to a distribution shift. Second, physical inconsistency leads to unrealistic artifacts. For instance, rigid objects may visually warp during contact or vanish entirely due to spatiotemporal inconsistencies.




While existing works attempt to mitigate this by incorporating failure trajectories, the effectiveness is limited as action coverage is infeasible~\cite{guo2025ctrl,ho20251x}. We argue that the bottleneck is fundamentally architectural. 
Most existing approaches adapt architectures originally designed for video generation (e.g., image-to-video models).
Since these backbones are not natively designed to take robotic actions as input, actions are merely injected as auxiliary conditions (e.g., via cross-attention or adaptive modulation like AdaLN) into the visual denoiser
~\cite{li2025worldeval,guo2025ctrl,quevedo2506worldgym}.
Given that these backbones are heavily pre-trained on massive video datasets, they inherit strong visual priors.
Consequently, action signals act as weak guidance and are frequently overridden by these dominant priors, leading to hallucinated success or spatiotemporal drift.

Motivated by this, we propose dWorldEval, a world model based on Masked Discrete Diffusion (MDD)~\cite{austin2021structured,sahoo2024simple,lou2023discrete,LLaDA,lavida,MMaDA}. 
Unlike pre-trained video backbones, dWorldEval is trained from scratch on robotic data, treating actions and visual observations as equivalent tokens to ensure action controllability.
Specifically, We map visual observations, language instructions, and action chunks into a unified token space, modeling them jointly via a self-attention backbone.
To enable reliable policy evaluation, we incorporate a sparse keyframe memory that maintains spatiotemporal consistency by mitigating long-horizon drift. Additionally, we introduce a discrete progress token to quantify task completion; by jointly predicting this token with future observations, the model automatically determines success when progress reaches 1.

In summary, we make three contributions:
\begin{itemize}
    \item We propose dWorldEval, a discrete-diffusion world model that significantly enhances action controllability, utilizing sparse keyframe memory to ensure spatiotemporal consistency.
    \item We jointly predict visual outcomes and a discrete progress token to enable automatic success detection.
    \item We conduct a systematic evaluation on LIBERO~\cite{liu2023libero}, RoboTwin~\cite{mu2025robotwin}, and real-world tasks. Extensive experiments confirm that dWorldEval achieves substantially better action controllability measured by our proposed action-sensitive $\Delta$-LPIPS metric. Furthermore, its estimated success rates correlate strongly with actual execution performance (Pearson $r \approx 0.9$), enabling accurate ranking of policies across capabilities.
\end{itemize}

\begin{figure*}[t]
  \centering
  \includegraphics[width=0.98\textwidth]{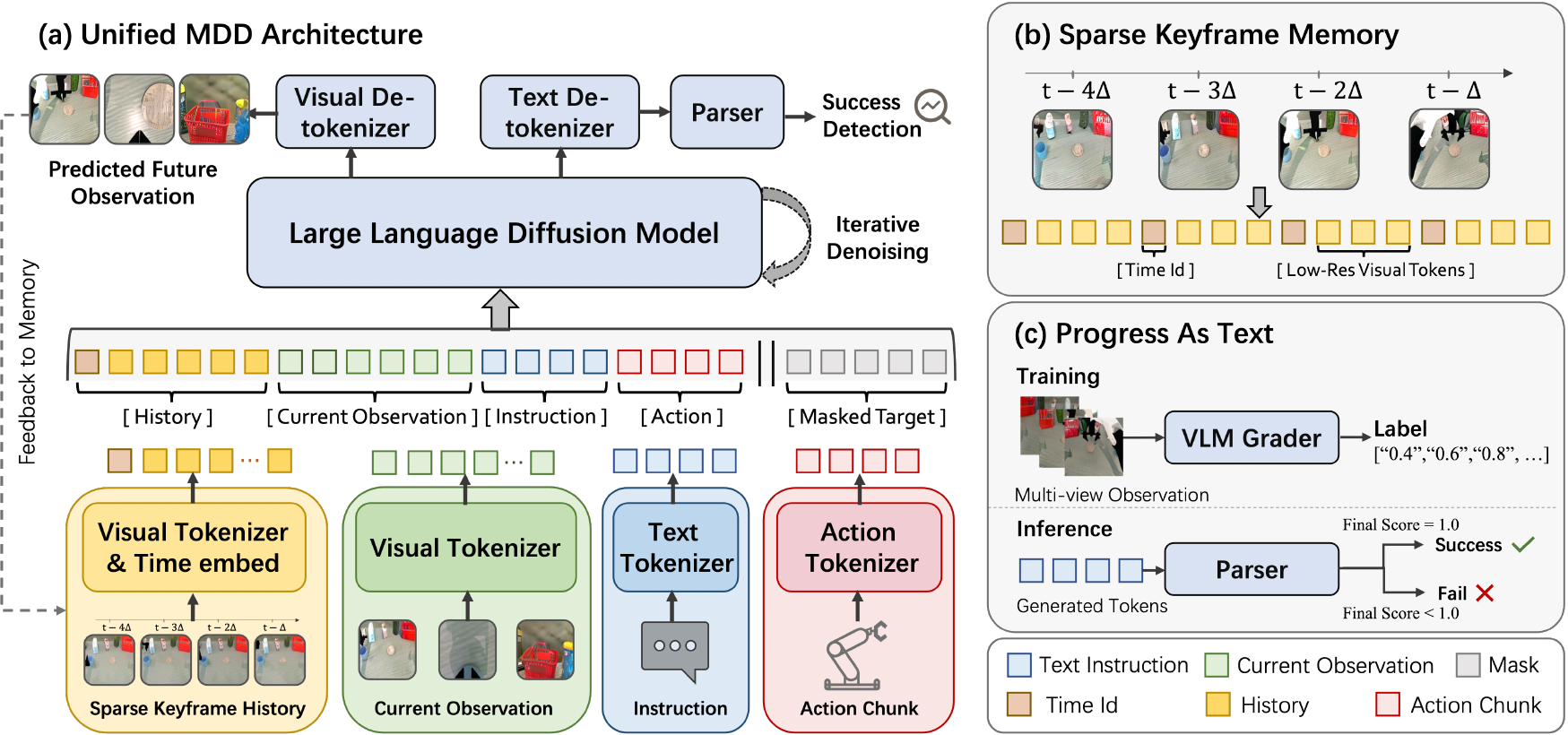}
  \caption{
    \textbf{Overview of dWorldEval.} 
    \textbf{(a) Unified architecture:} Diverse modalities are flattened into a single sequence, enabling the model to treat visual, control, and semantic tokens uniformly.
    \textbf{(b) Sparse keyframe memory:} A history of low-resolution keyframes is used to anchor global spatiotemporal consistency.
    \textbf{(c) Discrete progress token:} The model jointly predicts visual outcomes and discrete progress tokens to enable automatic success detection.
  }
  \label{fig:model}
\end{figure*}
\section{Related Work}
\paragraph{World models for policy evaluation.} 
While policy evaluation has traditionally depended on real-world rollouts~\cite{li24simpler, zhou2025autoeval} or physics-based simulators
~\cite{mees2022calvin, mujoco, mu2021maniskill, gu2023maniskill2, xiang2020sapien, makoviychuk2021isaac, mittal2023orbit, yu2020metaworld, mu2025robotwin, sferrazza2024humanoidbench, grotz2024peract2, liu2023libero}, world models offer a data-driven paradigm to scale policy assessment~\cite{li2025worldeval,guo2025ctrl,quevedo2506worldgym,team2025evaluating,1xWorldModel,ho20251x,huang2025enerverse,jiang2025enerverse,tseng2025scalable}. However, the utility of current video-based evaluators is severely limited by their lack of reliability. Fundamentally, these architectures treat actions as auxiliary conditions into a visually-dominated denoiser—e.g., via AdaLN modulation in WorldGym~\cite{quevedo2506worldgym} or cross-attention in Ctrl-World~\cite{guo2025ctrl} and WorldEval~\cite{li2025worldeval}. This design allows strong visual priors to override control signals, causing the model to frequently hallucinate transitions.
In contrast, dWorldEval integrates actions as primary tokens within a unified sequence, enabling the generated future states to be directly driven by control signals via self-attention.

\paragraph{Discrete diffusion in robotics.}
Discrete diffusion language models (DLMs) have emerged as strong alternatives to autoregressive LLMs, exhibiting competitive generative capabilities while enabling flexible sampling strategies and improved controllability~\cite{austin2021structured,sahoo2024simple,lou2023discrete,LLaDA}. Recent efforts extend DLM backbones to multimodal understanding and generation, e.g., LaViDa~\cite{lavida} and MMaDA~\cite{MMaDA}. In robotics, dVLA~\cite{wen2025dvla} and related works adapt discrete diffusion to policy learning, formulating action prediction as token inpainting over VLA-style inputs~\cite{wen2025dvla,liang2025discrete,wen2025LLaDA}.
In contrast to these policy learning approaches, we employ discrete diffusion to build a world model. Our unified token space enables joint prediction of visual observations and progress scores, eliminating the need for external VLMs or reward functions required by prior evaluators~\cite{li2025worldeval,quevedo2506worldgym,guo2025ctrl}.

\section{Methodology}
\subsection{Problem Formulation}
\label{sec:formulation}

\paragraph{World model formulation.}
We aim to learn a world model $\mathcal{W}_\theta$ that serves as a proxy for evaluating robotic policies.
Specifically, given a language instruction $l$, a current observation $o_t$, a history $h_t$, and a sequence of future actions $\mathbf{a}_t$, the model predicts the visual outcome $\hat{o}_{t+\Delta}$ and a task progress score $\hat{v}_{t+\Delta} \in [0,1]$ at horizon $\Delta$.
Essentially, $\mathcal{W}_\theta$ approximates the joint distribution of future visual dynamics and task completion: $(\hat{o}_{t+\Delta}, \hat{v}_{t+\Delta}) \sim \mathcal{W}_\theta(\cdot \mid o_t, \mathbf{a}_t, h_t, l)$.
To serve as a reliable evaluator, $\mathcal{W}_\theta$ must satisfy three properties:
(1) Action controllability: predictions must faithfully reflect the visual changes induced by the input actions;
(2) Spatiotemporal consistency: the model must preserve the spatial layout and consistency across long-horizon rollouts; and
(3) Discriminative task completion: generated future states must be semantically unambiguous to allow for accurate success detection.

\paragraph{Policy evaluation via imagination.}
We define an imagined rollout as a closed-loop interaction between a policy $\pi$ and the world model $\mathcal{W}_\theta$, yielding a generated trajectory $\tau = \{\hat{o}_0, \hat{o}_\Delta, \dots, \hat{o}_T\}$.
The policy's performance is evaluated by the imagined success rate $J = \frac{1}{N}\sum_{i=1}^N \mathcal{S}(\tau_i)$, where $\mathcal{S}(\cdot) \in \{0,1\}$ is a success indicator function that judges whether the generated outcome fulfills the task instruction.

\subsection{World Modeling via Discrete Diffusion}
\label{sec:architecture}

\subsubsection{Action Control via Unified Token Sequence}
\label{sec:action_tokenization}
To ensure action controllability, we integrate actions directly into a unified discrete token space, rather than treating them as auxiliary conditions. Specifically, we employ specialized tokenizers to map heterogeneous modalities into discrete codes: MAGVIT-v2~\cite{magvit-v2} for RGB observations, LLaDA~\cite{LLaDA} for language, and FAST~\cite{pertsch2025fast} for continuous action chunks $\mathbf{a}_t$. By serializing these codes into a single flattened sequence, the transformer can model the joint distribution of actions and observations.
Through self-attention, each visual token directly attends to action tokens, enabling fine-grained control at the token level. This allows that generated visual states accurately reflect the input actions.


\begin{figure*}[t]
  \centering
\includegraphics[width=0.95\textwidth]{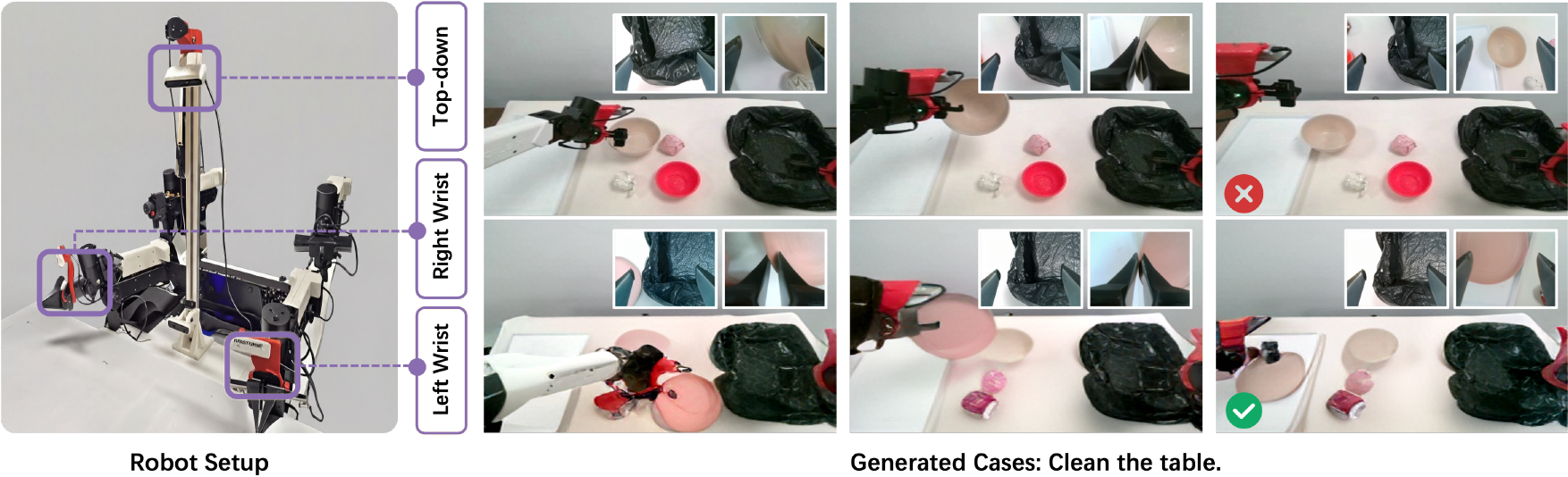}
    \caption{\textbf{Real-world setup and rollout visualization.} \textbf{Left:} The bimanual AgileX platform equipped with three synchronized cameras. \textbf{Right:} Result of a failure and a success rollout generated for the \textit{Clean the Table} task; the insets in the top corners of each frame display the synchronized wrist views.}
  \label{fig:real_setup}
\end{figure*}

\subsubsection{Sparse Keyframe Memory for Spatiotemporal Consistency}
\label{sec:training_history}

To maintain spatiotemporal consistency, we employ a sparse keyframe memory. The memory is updated via a sliding window that samples the most recent $K$ frames at a fixed stride $\Delta$, aligned with the action chunk size. These sampled keyframes are tokenized and concatenated with other tokens described in Sec.~\ref{sec:action_tokenization}. To explicitly preserve temporal order within this sequence, we encode absolute frame indices as text tokens and prepend them to the corresponding history keyframes~\cite{guo2025seed1}.
To optimize computational efficiency, we encode history frames at a reduced resolution, utilizing only the fixed global view (e.g., top-down). This setup provides sufficient context for the model while significantly reducing token usage. In contrast, the current observation retains all views at full resolution to capture the fine-grained object interactions required for precise generation.

\subsubsection{Discrete Progress Token for Automatic Success Detection}
\label{sec:training_score}
Prior methods decouple success detection from visual generation, incurring additional overhead and potential inconsistency. Instead, our unified token space enables joint generation of visual outcomes and progress scores within a unified latent space, thereby aligning the predicted score with the generated content.
During training, we define task-specific milestones and prompt SEED-1.5VL~\cite{guo2025seed1} with few-shot examples to estimate task completion progress from visual observations (see Appendix~\ref{app:progress_grading} for details). The resulting scores are converted into discrete tokens (e.g., ``1.0'') and appended to the target sequence.
At inference, the generated text tokens are decoded into a numeric score $\hat{v}_{t+\Delta}$, enabling automatic computation of success rate for policy ranking.

\subsection{Joint Visual-and-Progress Denoising}
\label{sec:backbone}
We employ Masked Discrete Diffusion (MDD) to learn the transition distribution $p(\mathbf{y}_{t+\Delta} \mid \mathbf{c}_t)$, where the context $\mathbf{c}_t$ remains unmasked while the target suffix is partially masked for reconstruction. Specifically, during training, we sample a diffusion level $\lambda \sim \mathcal{U}(0,1)$ and corrupt the target via a mask-based forward process $\tilde{\mathbf{y}}_{t+\Delta} \sim q_\lambda(\tilde{\mathbf{y}} \mid \mathbf{y}_{t+\Delta})$. The model is then optimized to reconstruct the clean tokens at the masked indices $\Omega_\lambda = \{j \mid \tilde y_j = \texttt{[MASK]}\}$ by minimizing the following weighted objective:
\begin{equation}
\mathcal{L}_{\text{WM}}
=
\mathbb{E}_{\tau,\lambda,\tilde{\mathbf{y}}}
\Bigg[
-\frac{1}{m(\lambda)}
\sum_{j\in\Omega_\lambda}
w_j \log p_\theta\!\left(
y_j \mid
\mathbf{c}_t,\tilde{\mathbf{y}}_{t+\Delta},\lambda
\right)
\Bigg],
\label{eq:loss_total}
\end{equation}
where $m(\lambda)$ denotes the masking probability at level $\lambda$, and $w_j$ serves as a modality-specific rebalancing weight. At inference, we use iterative parallel decoding to sample $\mathbf{y}_{t+\Delta}$ conditioned on the unified context $\mathbf{c}_t$. This process simultaneously generates the future visual state and its progress score.

\section{Experiments}
\label{sec:experiments}

In this section, we evaluate the generative capabilities of dWorldEval and its function as an robotic policy evaluator. 
Specifically, our investigation addresses the following research questions:
\begin{itemize}
    \item \textbf{(RQ1)}
    Does dWorldEval strictly adhere to control signals, faithfully rendering execution failures from suboptimal or OOD actions rather than hallucinating outcomes?

    \item \textbf{(RQ2)}
    Does the sparse keyframe memory
    (Sec.~\ref{sec:training_history}) 
    effectively prevent spatiotemporal drift, thereby maintaining long-horizon consistency required for evaluation?

    \item \textbf{(RQ3)}
    Can the proposed Progress-as-text mechanism serve as an accurate intrinsic success metric at inference time without external oracles?

    \item \textbf{(RQ4)}
    Is dWorldEval a reliable proxy for assessing diverse robot policies? Specifically, does the performance estimated via closed-loop rollouts strongly correlate with real execution and accurately rank policies across different architectures and training stages?

    \item \textbf{(RQ5)} 
    Can the proposed $\Delta$-LPIPS metric effectively quantify action controllability and serve as a reliable diagnostic indicator for policy evaluation?
\end{itemize}

\subsection{Experimental Setup}
\label{subsec:setup}

\textbf{Platforms and data.} 
We conduct evaluations across three diverse platforms, configuring the training data for each to support world modeling:

\textbf{(1) LIBERO}~\cite{liu2024libero}:
We utilize the LIBERO-Object, LIBERO-Spatial, LIBERO-Goal, and LIBERO-100 suites with synchronized third-person and wrist views.
To enable failure-aware scoring, we augment the 5.5k official expert demonstrations with 1k failed rollouts from suboptimal policies.

\textbf{(2) RoboTwin}~\cite{mu2025robotwin}:
We select the ARX arm configuration to evaluate contact-rich tableware manipulation, yielding 5.5k trajectories across 10 tasks (e.g., multi-object stacking and precise pick-and-place).

\textbf{(3) Real-world setup}:
We deploy a physical bimanual AgileX system (Figure~\ref{fig:real_setup}) with two 6-DoF arms and three synchronized RealSense 457 cameras.
The dataset totals 5.2k trajectories (including 1k human-collected failures) across five tasks: Bussing Table, Place Cup, Handover Block, Strike Block and Place Bottles.




\textbf{Baselines and target policies.} 
We benchmark dWorldEval against video diffusion baselines including WorldEval~\cite{li2025worldeval}, WorldGym~\cite{quevedo2506worldgym} and Ctrl-World~\cite{guo2025ctrl}. ensuring identical training data splits for fair comparison. 
And we assess its capabilities across varied policies: multiple training checkpoints of a base policy ($\pi_0$~\cite{[pi0}) on LIBERO, and heterogeneous architectures (e.g., DexVLA~\cite{wen2025dexvla}, Diffusion Policy~\cite{diffusion-policy}) across RoboTwin and real-world environments.

\begin{figure*}[t]
    \centering
    \includegraphics[width=0.9\textwidth]{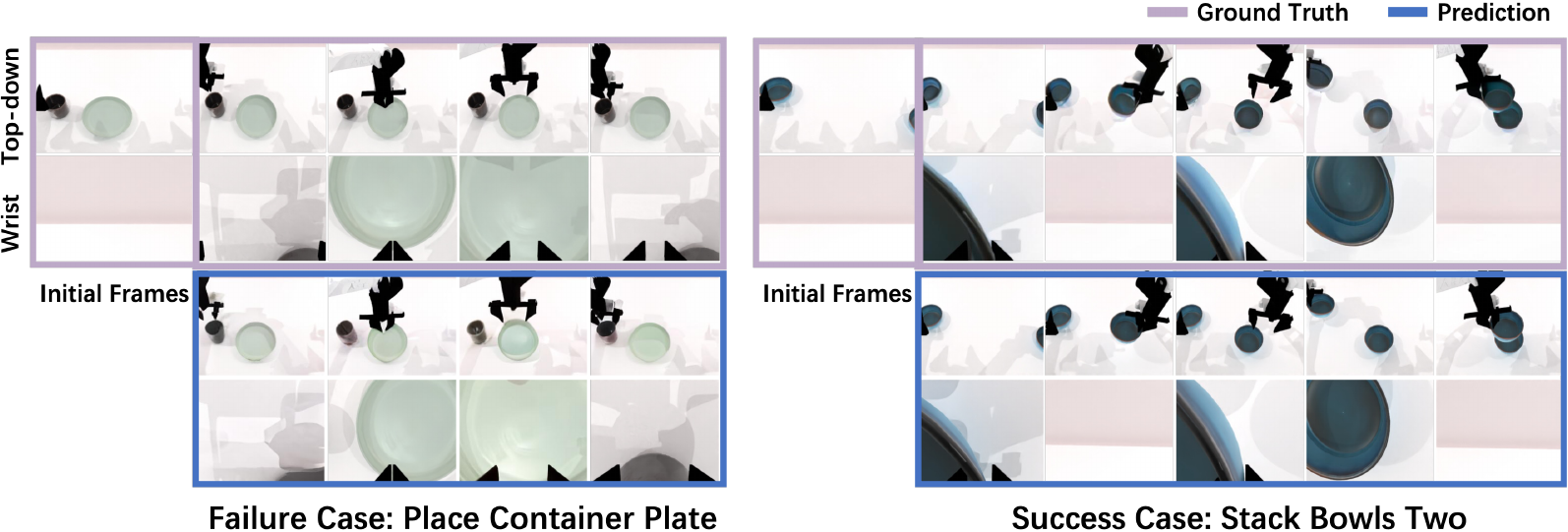}
    \caption{\textbf{Visualization of ground-truth vs.\ generated multi-view rollouts on RoboTwin}~\cite{mu2025robotwin}.
    \textbf{Left/Right}: suboptimal vs.\ successful execution.
    \textbf{Top/Bottom}: ground-truth simulator rollout vs.\ dWorldEval prediction, conditioned on identical action sequences. 
    Each sequence displays the initial state followed by synchronized future frames.}
    \label{fig:fail_success_robotwin}
\end{figure*}

\textbf{Implementation details.} 
All models predict multi-view outcomes at $256^2$ resolution, conditioned on a sparse history of $K=4$ keyframes ($128^2$). 
Regarding the loss weights in Eq.~\ref{eq:loss_total}, we set $w_j = 2$ for progress tokens and $w_j = 1$ for visual tokens.
The prediction horizon $\Delta$ aligns with the action chunk length (selected from $[2, 8]$). 
During inference, we employ 16-step iterative parallel decoding.

\begin{figure*}[t]
    \centering
    \includegraphics[width=\textwidth]{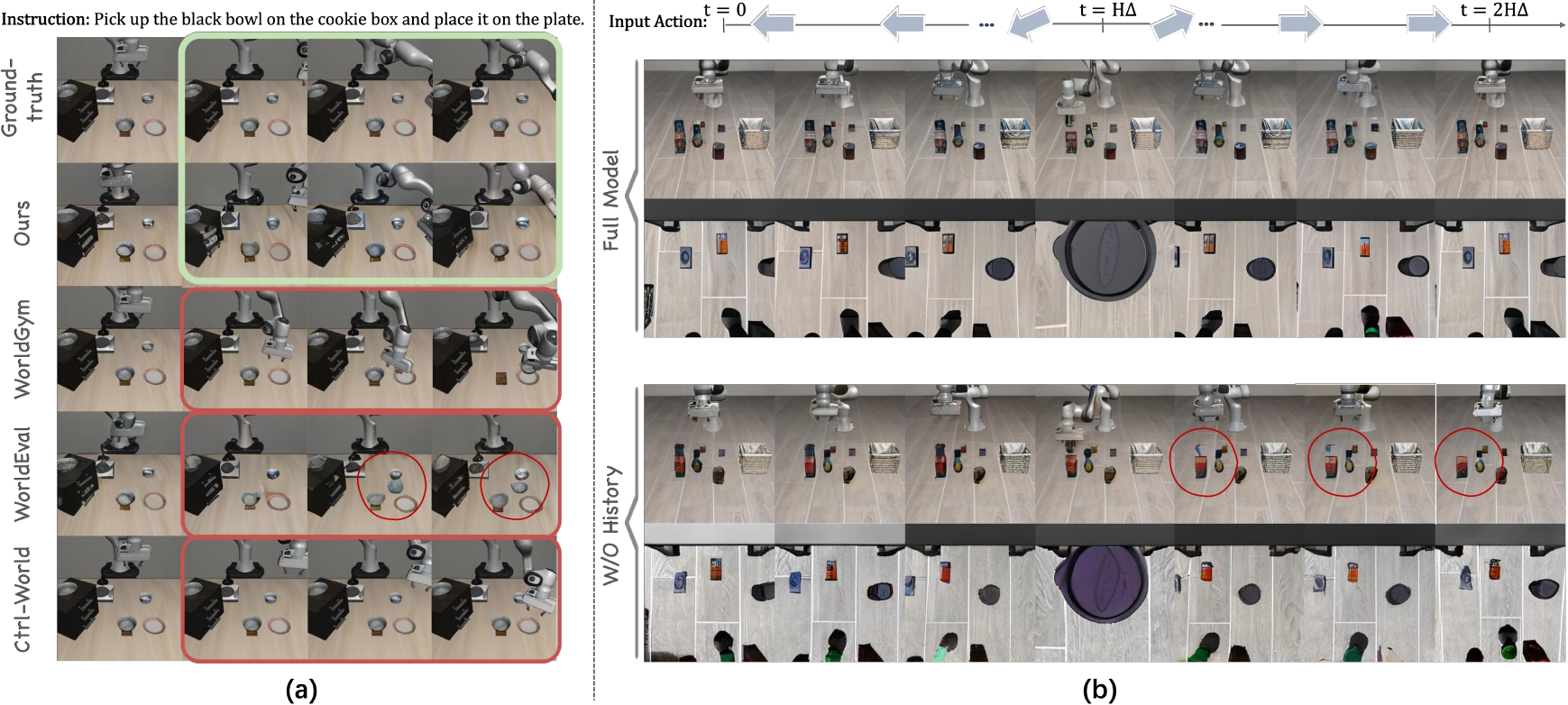}
    \caption{\textbf{(a) Action-controllability under suboptimal inputs.}
    We condition all methods on an unseen action sequence and compare their rollouts.
    While WorldGym~\cite{li2025worldeval} self-corrects the missed grasp into a successful pickup and WorldEval~\cite{li2025worldeval} hallucinates non-existent objects, Ctrl-World~\cite{guo2025ctrl} fails to align with the input actions.
    In contrast, dWorldEval faithfully reproduces the failure.
    \textbf{(b) Long-horizon round-trip consistency.}
    We apply a reversible action trajectory: forward actions up to $t=H\Delta$, followed by the corresponding inverse actions to $t=2H\Delta$.
    The full model returns close to the initial observation at $t=0$, whereas removing history causes accumulated drift over the round trip.}
    \label{fig:consistency}
\end{figure*}

\begin{figure*}[t]
    \centering
    \includegraphics[width=0.98\textwidth]{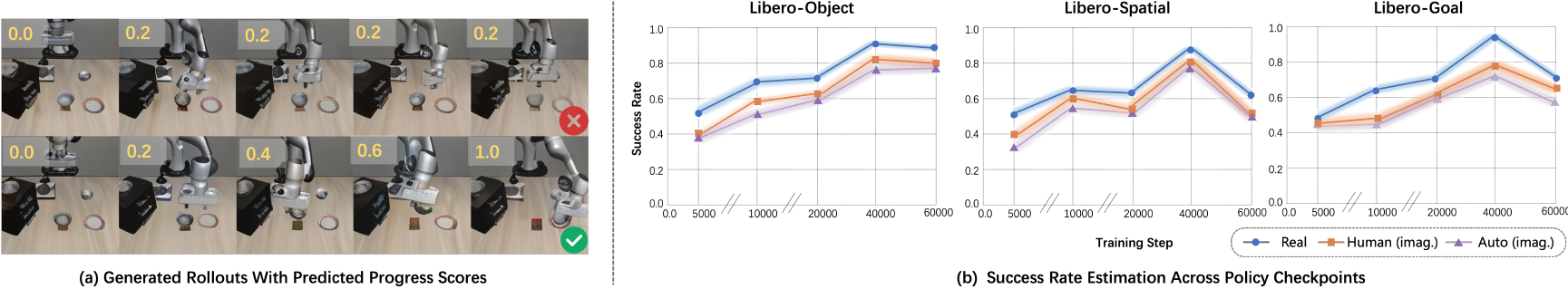}
    \caption{\textbf{Joint generation enables automatic policy scoring.}
    \textbf{(a)} At each rollout step, the world model jointly predicts the future observation $\hat{o}_{t+\Delta}$ and a scalar progress score $\hat{v}_{t+\Delta}\!\in[0,1]$, which finally indicates task success or failure.
    \textbf{(b)} Success rate estimates across checkpoints of a base policy $\pi_0$~\cite{[pi0} on three LIBERO~\cite{liu2024libero} suites.
    We compare ground-truth execution (\emph{Real}) against evaluations on generated rollouts: human judgment based on the generated observations (\emph{Human}) and automatic evaluation determined by the model's predicted progress score (\emph{Auto}).}
    \label{fig:score_curve}
\end{figure*}

\begin{figure*}[t]
    \centering
    \includegraphics[width=0.98\textwidth]{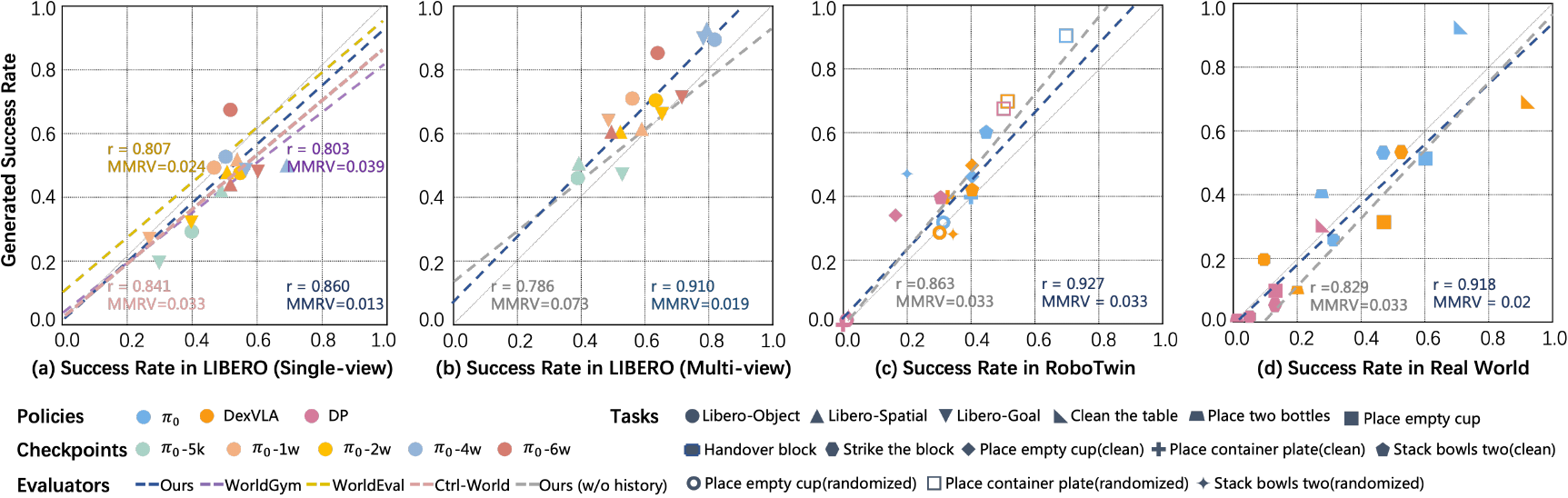}
    \caption{\textbf{Correlation between real-execution and world-model success rates.}
    Scatter plots compare real success rates (x-axis) against generated estimates (y-axis), reporting Pearson correlation $r$ and rank-violation \textsc{MMRV} (lower is better).
    \textbf{(a)} Comparison with video diffusion baselines (WorldEval~\cite{li2025worldeval}, WorldGym~\cite{quevedo2506worldgym} and Ctrl-World~\cite{guo2025ctrl}) on LIBERO~\cite{liu2024libero} (Single-view).
    \textbf{(b-d)} Ablation studies comparing dWorldEval against its \textit{w/o-history} variant across diverse settings:
    \textbf{(b)} LIBERO (Multi-view),
    \textbf{(c)} Robotwin~\cite{mu2025robotwin} with heterogeneous policies ($\pi_0$~\cite{[pi0}, DexVLA~\cite{wen2025dexvla}, DP~\cite{diffusion-policy}), and
    \textbf{(d)} Real-world tasks.
    dWorldEval consistently achieves superior correlation and ranking accuracy across all domains.}
    \label{fig:policy_corr}
\end{figure*}

\subsection{Evaluating the World Model}
\label{sec:wm_eval}

\subsubsection{Evaluating Action Controllability}
\label{sec:action_controllability_exp}

\paragraph{Evaluation protocol.}
To assess action controllability (RQ1), we employ a protocol on a test set constructed from two distinct interaction types:
(1) Expert success ($\mathcal{D}_{\text{succ}}$): successful trajectories collected from a fully trained policy;
(2) Suboptimal failure ($\mathcal{D}_{\text{fail}}$): failure rollouts collected from undertrained checkpoints.
We perform generation conditioned on ground-truth action sequences and evaluate on the shared third-person view.

\textbf{Evaluation metric.}
Standard LPIPS captures only static appearance. To explicitly evaluate action controllability, we introduce \textbf{$\Delta$-LPIPS}, which measures the perceptual fidelity of state transitions rather than absolute states. For a fixed stride $\Delta$, we compute difference images $\Delta o_t = o_{t+\Delta} - o_t$ (analogously for predictions $\Delta \hat{o}_t$) and define:
\begin{equation}
  \Delta\mathrm{LPIPS} = \mathbb{E}_{t}\Big[d_{\mathrm{lpips}}\big(\mathrm{norm}(\Delta \hat{o}_t), \mathrm{norm}(\Delta o_t)\big)\Big],
\end{equation}
where $\mathrm{norm}(\cdot)$ denotes per-sample RMS normalization for stability. We use $\Delta$LPIPS as our primary indicator for action-conditioned dynamic fidelity (see Sec.~\ref{sec:more_experiments} for further validation of its diagnostic correlation).

\paragraph{Experimental results.}
This visual fidelity is quantified in Tab.~\ref{tab:wm_lpips}, which exposes a critical divergence: baselines suffer severe degradation on the Failure subset (e.g., WorldGym $\Delta$LPIPS spikes from 0.347 to 0.650), whereas dWorldEval maintains consistent performance. This quantitative degradation manifests visually in Fig.~\ref{fig:consistency}(a), where baselines frequently fail to adhere to the input actions.
To verify these results stem from action controllability, we find that randomly shuffling actions (App.~\ref{app:action_shuffle}) significantly degrades $\Delta$LPIPS, proving our model is sensitive to the input actions.

\begin{table}[t]
    \centering
    
    \begin{minipage}[t]{0.48\textwidth}
        \centering
        \small
        \setlength{\tabcolsep}{8pt} 
        \caption{\textbf{Action controllability on LIBERO}~\cite{liu2024libero}. Comparison of LPIPS vs. our dynamic-aware $\Delta$LPIPS on Expert ($\mathcal{D}_{\text{succ}}$) and Failure ($\mathcal{D}_{\text{fail}}$) subsets. Baselines degrade on failure data, while ours remains robust.}
        \label{tab:wm_lpips}
        \begin{sc}
            \begin{tabular}{lccc} 
                \toprule
                \multirow{2}{*}{Method} & LPIPS & \multicolumn{2}{c}{$\Delta$LPIPS ($\downarrow$)} \\
                \cmidrule(lr){3-4} 
                 & ($\downarrow$) & Expert & Failure \\
                \midrule
                WorldEval & 0.262 & 0.423 & 0.701 \\
                WorldGym  & 0.218 & 0.347 & 0.650 \\
                Ctrl-World  & 0.220 & 0.334 & 0.416 \\
                \midrule
                \textbf{Ours} & \textbf{0.215} & \textbf{0.315} & \textbf{0.352} \\
                \bottomrule
            \end{tabular}
        \end{sc}
    \end{minipage}%
    \hfill 
    \begin{minipage}[t]{0.48\textwidth}
        \centering
        \small
        \setlength{\tabcolsep}{2.4pt} 
        \caption{\textbf{Long-horizon consistency analysis.} We report the round-trip LPIPS ($\downarrow$) error, averaged across all synchronized views, over varying horizons $H$. Without memory, the model suffers from progressive drift; with memory, spatiotemporal consistency is effectively preserved even at $H=20$.}
        \label{tab:h_scan_consistency}
        \begin{sc}
            \begin{tabular}{lcccc}
            \toprule
            \multirow{2}{*}{Method} & \multicolumn{4}{c}{Action Horizon ($H$)} \\
            \cmidrule(lr){2-5}
             & 5 & 10 & 15 & 20 \\
            \midrule
            Ours (W/O Memory) & 0.177 & 0.186 & 0.302 & 0.411 \\
            \textbf{Ours (Full)} & \textbf{0.130} & \textbf{0.145} & \textbf{0.193} & \textbf{0.243} \\
            \bottomrule
            \end{tabular}
        \end{sc}
    \end{minipage}
    
\end{table}

\subsubsection{Evaluating Spatiotemporal Consistency}
\label{sec:exp_consistency}

\paragraph{Evaluation protocol and metrics.}
To assess long-horizon stability (RQ2), we employ a variable-horizon round-trip protocol. We perform rollouts of length $2H$ by appending inverse actions to trajectories of varying lengths $H \in \{5, 10, 15, 20\}$. Consistency is measured via LPIPS  between the initial $o_t$ and final $\hat{o}_{t+2H}$.
We focus on memory ablation here. Baselines are detailed in Appendix~\ref{app:baseline_consistency}, as their inability to strictly follow actions  renders the round-trip metric unreliable for direct comparison.

\paragraph{Experimental results.}
Fig.~\ref{fig:consistency}(b) visualizes the cumulative drift isolated by removing memory.
This is quantified in Tab.~\ref{tab:h_scan_consistency}: while ablation errors accumulate as $H$ extends to 20, the full model maintains high fidelity (LPIPS $0.21$), confirming that keyframe memory effectively bounds long-term drift. This stability is vital: Sec.~\ref{sec:policy_eval} shows that drift leads to false negatives, severing the correlation with real-world performance.

\subsubsection{Joint subgoal-and-progress prediction enables automatic visual grading}
\label{sec:joint_progress}

\textbf{Evaluation protocol.}
We leverage the failure-augmented training regime (Sec.~\ref{subsec:setup}) to validate whether the learned progress tokens can serve as an intrinsic metric (RQ3). 
We evaluate multiple checkpoints of a base policy $\pi_0$ across four LIBERO suites and compare three success estimators:
(1) \textbf{Real}: Ground-truth success rates measured by real execution;
(2) \textbf{Human (imag.)}: Manual grading of the final generated frame;
(3) \textbf{Auto (imag.)}: Our proposed automated metric. For each imagined rollout, we examine the predicted progress token $\hat{v}_{T}$ at the final time step. The rollout is classified as successful only if this terminal score equals 1.

\paragraph{Experimental results.} As shown in Fig.~\ref{fig:score_curve}(a), the progress score is discriminative, exhibiting sharp transitions upon task completion. Crucially, Fig.~\ref{fig:score_curve}(b) demonstrates that Auto (imag.) closely tracks real execution, capturing even non-monotonic fluctuations (e.g., performance dips in later checkpoints).
This alignment closely matches human judgment. It confirms that dWorldEval bases its score on the actual visual changes, rather than simply assuming progress increases over time. This ensures the reliability of our automatic evaluation.

\subsection{World-Model as a Reliable Proxy for Policy Evaluation}
\label{sec:policy_eval}

For the ablation without memory, the generated video often suffers from severe drift, making the predicted progress token unreliable. To ensure a fair comparison, we ignore the progress and judge success  based on the generated image.

\textbf{Comparison with baselines on LIBERO.}
Fig.~\ref{fig:policy_corr}(a) evaluates policy ranking accuracy on LIBERO tasks under the single-view setting.
Existing baselines (WorldGym, WorldEval and Ctrl-World) exhibit weaker correlations and higher rank violations (MMRV up to 0.039) due to insufficient action controllability.
In contrast, dWorldEval achieves a strong linear correlation with minimal rank violation (MMRV=0.013), confirming that action-faithful generation is a prerequisite for reliable evaluation.

\textbf{Ranking heterogeneous policies across domains.} 
We further assess robustness across multi-view settings, heterogeneous architectures (RoboTwin), and physical environments (Real World).
As shown in Fig.~\ref{fig:policy_corr}(b-d), the \textit{w/o-history} ablation suffers from significant performance degradation, particularly in the multi-view setting ($r$ drops to 0.786).
In contrast, by preserving spatiotemporal consistency, dWorldEval achieves high correlations with actual execution success rates across LIBERO multi-view ($r=0.910$), Robotwin ($r=0.927$), and real-world ($r=0.918$) tasks.

\begin{table}[t]
    \centering
    
    \begin{minipage}[t]{0.5\textwidth}
        \vspace{0pt} 
        \centering
        \small
        \setlength{\tabcolsep}{3pt} 
        \caption{\textbf{Universal multi-view fidelity.} Quantitative evaluation across diverse platforms (LIBERO~\cite{liu2024libero}, RoboTwin~\cite{mu2025robotwin}, Real-Robot) and synchronized viewpoints. dWorldEval maintains consistent high fidelity (low $\Delta$LPIPS) on both simulation and real-world data.}
        \label{tab:uni_fidelity}
        \begin{sc}
            \begin{tabular}{ll|cc}
                \toprule
                Dataset & Viewpoint & LPIPS ($\downarrow$) & $\Delta$LPIPS ($\downarrow$) \\
                \midrule
                \multirow{2}{*}{LIBERO} & 3rd-Person & 0.215 & 0.324 \\
                                        & Wrist & 0.192 & 0.303 \\
                \midrule
                \multirow{2}{*}{RoboTwin} & Top-Down & 0.199 & 0.317 \\
                                          & Wrist & 0.220 & 0.345 \\
                \midrule
                \multirow{2}{*}{Real World} & Top-Down & 0.272 & 0.365 \\
                                            & Wrist & 0.254 & 0.336 \\
                \bottomrule
            \end{tabular}
        \end{sc}
    \end{minipage}%
    \hfill
    \begin{minipage}[t]{0.46\textwidth}
        \vspace{0pt} 
        \centering
        \small 
        \renewcommand{\arraystretch}{1} 
        \setlength{\tabcolsep}{6pt}
        \caption{\textbf{Full consistency comparison.} We report the round-trip LPIPS ($\downarrow$) error averaged over varying horizons $H$. Comparisons with WorldEval~\cite{li2025worldeval}, WorldGym~\cite{quevedo2506worldgym} and Ctrl-World~\cite{guo2025ctrl}, demonstrate that our model maintains superior spatiotemporal consistency, whereas baselines suffer from significant drift.}
        \label{tab:baseline_consistency_full}
        \begin{sc}
            \begin{tabular}{lcccc}
                \toprule
                \multirow{2}{*}{Model} & \multicolumn{4}{c}{Action Horizon ($H$)} \\
                \cmidrule(lr){2-5}
                 & 5 & 10 & 15 & 20 \\
                \midrule
                WorldEval    & 0.216 & 0.320 & 0.467 & 0.531 \\
                WorldGym   & 0.191 & 0.224 & 0.308 & 0.482 \\
                Ctrl-World   & 0.149 & 0.233 & 0.296 & 0.370 \\
                \textbf{Ours (Full)} & \textbf{0.130} & \textbf{0.145} & \textbf{0.193} & \textbf{0.243} \\
                \bottomrule
            \end{tabular}
        \end{sc}
    \end{minipage}
\end{table}

\subsection{More Experiments and Ablation Study}
\label{sec:more_experiments}

\textbf{Universal fidelity across diverse platforms.} 
We provide a comprehensive assessment of visual fidelity across diverse platforms.
As detailed in Tab.~\ref{tab:uni_fidelity}, dWorldEval maintains consistently low $\Delta$LPIPS scores ($\approx 0.31$-$0.36$) across varying camera configurations and robot morphologies. 
Notably, the performance in the real-world setup remains comparable to simulation results. This consistency suggests that our unified tokenization is robust against domain gaps, maintaining high fidelity across diverse environments.

\textbf{Comparison of long-horizon consistency.} While Sec.~\ref{sec:exp_consistency} focused on memory ablation, here we benchmark against video diffusion baselines.
As detailed in Tab.~\ref{tab:baseline_consistency_full}, baselines suffer from severe inconsistency that worsens as the horizon extends.
We interpret this degradation as a compound failure: since baselines frequently ignore control signals (as established in Sec.~\ref{sec:action_controllability_exp}), their high LPIPS errors stem not only from spatiotemporal drift but also from insufficient action controllability, where the model fails to follow the action sequence.
In contrast, dWorldEval maintains high spatiotemporal consistency. Qualitative visualizations of these failure modes are provided in Appendix~\ref{app:baseline_consistency}.

\section{Conclusion}
We presented dWorldEval, a discrete diffusion world model designed to overcome the limitations of existing methods in reliability. 
Our approach unifies action tokens with sparse keyframe memory to generate consistent long-horizon rollouts, quantified by our proposed $\Delta$LPIPS metric. 
Extensive experiments confirm that dWorldEval significantly enhances controllability, with predicted success rates that closely track real-world execution. 
This alignment enables accurate policy ranking across diverse architectures, bringing scalable evaluation closer to practice.

\clearpage
\bibliographystyle{plainnat}
\bibliography{main}

\clearpage

\begin{appendices}
    
\section{More on Experimental Setup}
\label{app:setup}

In this section, we provide further details regarding the task definitions, data collection pipelines, and model hyperparameters used in our experiments.

\subsection{Detailed Task Descriptions}

\begin{figure}[h]
    \centering
    \includegraphics[width=0.9\textwidth]{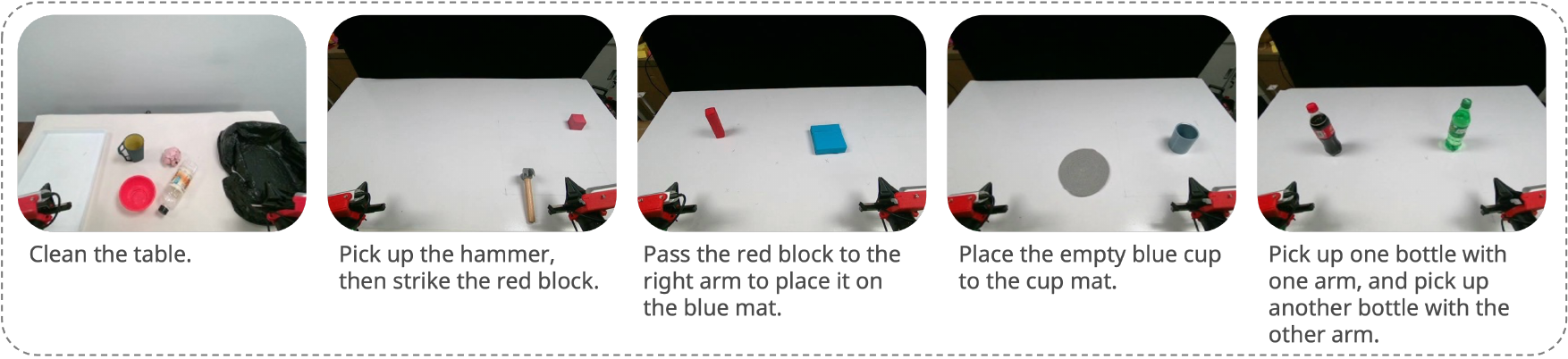}
    \caption{
      \textbf{Real-World Evaluation Tasks.} 
      We visualize the initial scene observations and corresponding language instructions for the five bimanual manipulation tasks evaluated on the AgileX platform.
    }
    \label{fig:real_tasks}
\end{figure}

We evaluate dWorldEval across three domains: Real-World AgileX, RoboTwin~\cite{mu2025robotwin}, and ~\cite{liu2024libero}. Below we describe the specific task designs and their associated language instructions.

\textbf{Real-World Tasks}
We collected data for five distinct tasks using a bimanual AgileX system. Initial states and corresponding instructions are visualized in Fig.~\ref{fig:real_tasks}.

\begin{itemize}
    \item \textbf{Bussing Table.} A long-horizon task requiring the robot to classify and sort multiple objects (tableware, trash) into trays or bins. \\
    \textit{Instruction: ``Clean the table.''}
    
    \item \textbf{Place Cup.} Precision pick-and-place where the robot must align a cup onto a specific target mat. \\
    \textit{Instruction: ``Place the empty blue cup to the cup mat.''}
    
    \item \textbf{Handover Block.} A bimanual coordination task involving passing a block from the left arm to the right arm before placement. \\
    \textit{Instruction: ``Pass the red block to the right arm to place it on the blue mat.''}
    
    \item \textbf{Strike Block.} Dynamic tool manipulation that requires grasping a hammer to strike a target block. \\
    \textit{Instruction: ``Pick up the hammer, then strike the red block.''}

    \item \textbf{Dual Bottle Pick.} A synchronization task requiring the robot to simultaneously grasp and lift two bottles positioned in front of it. \\
    \textit{Instruction: ``Pick up one bottle with one arm, and pick up another bottle with the other arm.''}
\end{itemize}

\textbf{Simulation Tasks}
\begin{itemize}
    \item \textbf{RoboTwin Stacking \& Placement.} On the RoboTwin benchmark, we utilize the ARX dual-arm configuration to evaluate 10 diverse contact-rich tasks. 
    We categorize these into: 
    (1) \textbf{Precision Stacking}, which requires vertical stability and geometry alignment (e.g., Stack Blocks (Three), Stack Bowls (Two \& Three)); and 
    (2) \textbf{Constrained Placement}, which involves inserting objects into specific receptacles, such as placing Bread (into Basket/Skillet), Cans (into Basket/Pot/Plastic Box), Plate (into Container), and an Empty Cup. 
    These scenarios involve complex inter-object occlusions and fine-grained physics that are challenging for video generation models.
    
    \item \textbf{LIBERO Suites.} We utilize four task suites from the LIBERO benchmark: LIBERO-Object, LIBERO-Spatial, LIBERO-Goal, and LIBERO-100. These tasks involve standard tabletop manipulation skills like opening drawers, moving objects around obstacles, and arranging items based on spatial relations.
\end{itemize}

\subsection{Model Implementation Details}
dWorldEval is initialized from MMaDAVLA-8B, a bidirectional transformer with 32 layers, 32 attention heads, and a hidden dimension of 4096. 
The model is conditioned on a sparse history of $K=4$ keyframes ($256 \times 256$). 
We use a prediction horizon $\Delta$ consistent with the action chunk size (randomly sampled from {$[2, 8]$}.
The loss function balances visual reconstruction and progress prediction with weights $w_{\text{score}}=2$ and $w_{\text{vis}}=1$. 
The model was trained for 15 epoch on 8 H800 GPUs using the AdamW optimizer with a learning rate of 5e-5.
We employ a global batch size of 128 (using gradient accumulation).We report success rates averaged over 20 episodes per task for simulation benchmarks and 30 episodes for real-world experiments.
Evaluating a full trajectory takes approximately 30--90 seconds (1.5s/frame) on a single H800 GPU.

\section{Verifying Causal Dependency via Action Shuffling}
\label{app:action_shuffle}

A faithful action-conditioned world model must genuinely depend on the provided future action chunk. 
If we deliberately destroy the alignment between actions and outcomes, action-controllability indicators should degrade accordingly.

\paragraph{Experimental Protocol.}
To test this dependency, we disrupt the causal link between actions and future observations while preserving their marginal distributions. 
Given a test batch $\{(o^i_t, h^i, l^i, a^i, o^i_{t+\Delta})\}_{i=1}^B$, we replace the true action $a^i$ with a randomly assigned $\tilde{a}^i \leftarrow a^{\pi(i)}$ via a batch-wise permutation $\pi$ (enforcing $\pi(i)\neq i$). 
We then roll out the world model under two settings:
(i) \textbf{Aligned}: $\hat{o}^i_{t+\Delta} \sim \mathcal{W}_\theta(o^i_t, h^i, l^i, a^i)$, and
(ii) \textbf{Shuffled}: $\tilde{o}^i_{t+\Delta} \sim \mathcal{W}_\theta(o^i_t, h^i, l^i, \tilde{a}^i)$.
Furthermore, to analyze the continuous impact of misalignment and validate $\Delta$-LPIPS as a diagnostic metric, we introduce an inference-time corruption: swapping action chunks with a tunable probability $p$. 
All evaluation metrics are computed identically under these settings on the same test set used in Sec.~\ref{sec:action_controllability_exp}.

\paragraph{Results.}
We first examine the complete disruption of action alignment ($p=1$). 
As shown in Tab.~\ref{tab:action_shuffle}, shuffling actions consistently degrades the action-controllability indicators. 
This confirms that our metrics are highly sensitive to the action input and that the learned dynamics are not simply relying on static appearance priors.

\begin{figure}[t]
    \centering
    \includegraphics[width=0.45\linewidth]{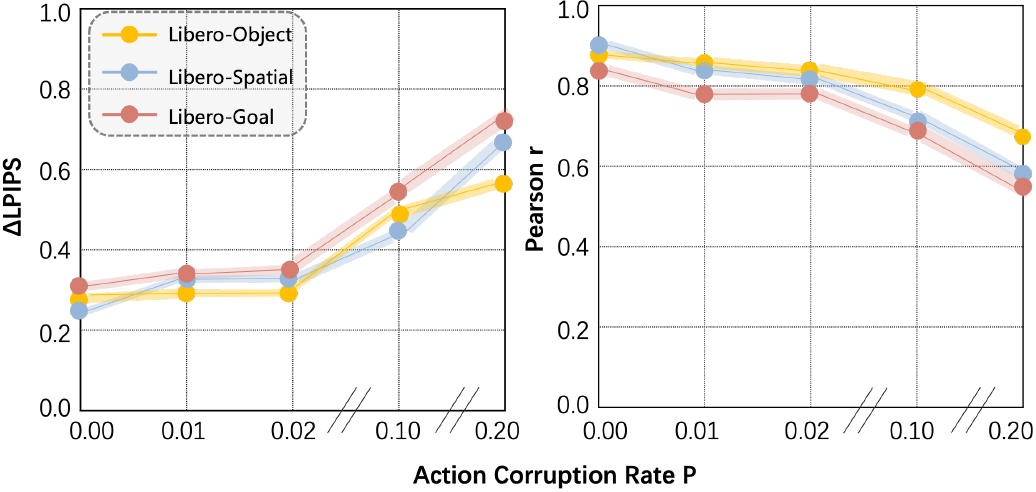}
    \caption{\textbf{Action corruption tests.} 
    Ground-truth action chunks are replaced by chunks from other episodes with probability $p$. 
    \textbf{Left:} $\Delta$LPIPS increases with $p$, indicating degraded controllability. 
    \textbf{Right:} Pearson correlation between real and estimated success rates drops accordingly.}
    \label{fig:ablate_action_corruption}
\end{figure}

Building upon this, the probabilistic corruption ($0 \le p \le 1$) reveals a strong correlation between action alignment and evaluation reliability. 
As illustrated in Fig.~\ref{fig:ablate_action_corruption}, increasing the swap probability $p$ worsens $\Delta$-LPIPS, accompanied by a precipitous drop in the ranking correlation with real success rates. 
This diagnostic demonstrates that accurate policy ranking is only achievable in the low-$\Delta$-LPIPS regime, reaffirming that strict adherence to input actions is a prerequisite for reliable evaluation.

\begin{table}[t]
\centering
\caption{\textbf{Action-shuffle sanity check.} We permute future action chunks across samples while keeping history and language fixed. Shuffling breaks action--outcome alignment and degrades action-controllability indicators.}
\label{tab:action_shuffle}
\begin{tabular}{lccc}
\toprule
Conditioning & LPIPS $\downarrow$ & $\Delta$LPIPS $\downarrow$ \\
\midrule
Aligned actions & \textbf{0.215} & \textbf{0.324} &  \\
Shuffled actions & 0.461 & 0.697  \\
\bottomrule
\end{tabular}
\end{table}


\section{VLM Supervision Details}
\label{app:progress_grading}

We utilize an off-the-shelf VLM (SEED-1.5VL~\cite{guo2025seed1}) to generate ground-truth progress scores. Unlike standard zero-shot evaluation, we employ a few-shot In-Context Learning (ICL) strategy to align the model's scoring distribution with human intuition.
For each query, we construct a prompt containing:
\begin{enumerate}
    \item A detailed task definition and rigid scoring rules.
    \item Three anchor examples with pre-labeled scores (e.g., 0.2, 0.4, and 0.6) to demonstrate intermediate states.
    \item A batch of query frames (typically 10 frames) to be evaluated independently.
\end{enumerate}
This batch-processing approach significantly stabilizes the output and enforces strict adherence to the discrete scoring criteria.

\subsection{Prompt Template}
For the Libero-Object~\cite{liu2024libero} suite, which primarily involves pick-and-place manipulation, we redefine the scoring criteria to reflect the sequential phases of the action. We employ an few-shot ICL strategy to help the VLM distinguish subtle state changes.

The system instruction maps the continuous manipulation process into discrete progress steps:

\begin{quote}
\small
\textbf{System Instruction:} \\
You are an expert roboticist evaluator. Your task is to judge the completion progress of a robot performing a specific manipulation task (e.g., ``pick up the bbq sauce and place it in the basket"). \\
\textbf{Task Instruction:} \{TASK\_INSTRUCTION\} \\
\textbf{Scoring Rules (Progress Phases):} 
The task progress is discretized into the following stages. Please choose the score that best matches the current visual state:
\begin{itemize}
    \item \textbf{0.0}: \textit{Idle / Start}. The robot has not yet interacted with the target object.
    \item \textbf{0.2}: \textit{Approach / Contact}. The gripper is positioned near the target object or has just made contact, but the object is not yet lifted.
    \item \textbf{0.4}: \textit{Lifted}. The object is successfully grasped and lifted off the surface.
    \item \textbf{0.6}: \textit{In Transit}. The robot is moving the object towards the target area (mid-trajectory).
    \item \textbf{0.8}: \textit{Pre-Placement}. The object is aligned with or hovering just above the target zone, ready for release.
    \item \textbf{1.0}: \textit{Success}. The object is stably placed in the target configuration, and the gripper has released it (or the task is fully complete).
\end{itemize}
\textbf{Important note:} The input frames may appear in random order. You must evaluate each frame independently, strictly based on the visible state in that frame. Do not infer progress from previous frames. \\
\textbf{Valid outputs:} 0, 0.2, 0.4, 0.6, 0.8, 1.0 \\
\textbf{Output Format:} Return only a list of numbers (e.g., [0.2, 0, 0.6, 1.0]).
In-Context Examples: \\
We provide 8 distinct examples covering the full range of motion to anchor the scoring: \\
\textit{[Image 1]} (Start State) $\rightarrow$ Text: "Example 1 score: 0.0" \\
\textit{[Image 2]} (Approaching) $\rightarrow$ Text: "Example 2 score: 0.2" \\
\textit{[Image 3]} (Just Lifted) $\rightarrow$ Text: "Example 3 score: 0.4" \\
\textit{[Image 4]} (Mid-Air Moving) $\rightarrow$ Text: "Example 4 score: 0.6" \\
\textit{[Image 5]} (Approaching Target) $\rightarrow$ Text: "Example 5 score: 0.6" \\
\textit{[Image 6]} (Aligned/Hovering) $\rightarrow$ Text: "Example 6 score: 0.8" \\
\textit{[Image 7]} (Just Released) $\rightarrow$ Text: "Example 7 score: 1.0" \\
\textit{[Image 8]} (Retracting/Done) $\rightarrow$ Text: "Example 8 score: 1.0" \\
Text: "Now evaluate these frames:"
\end{quote}

\begin{figure*}[t]
    \centering
    \includegraphics[width=0.85\textwidth]{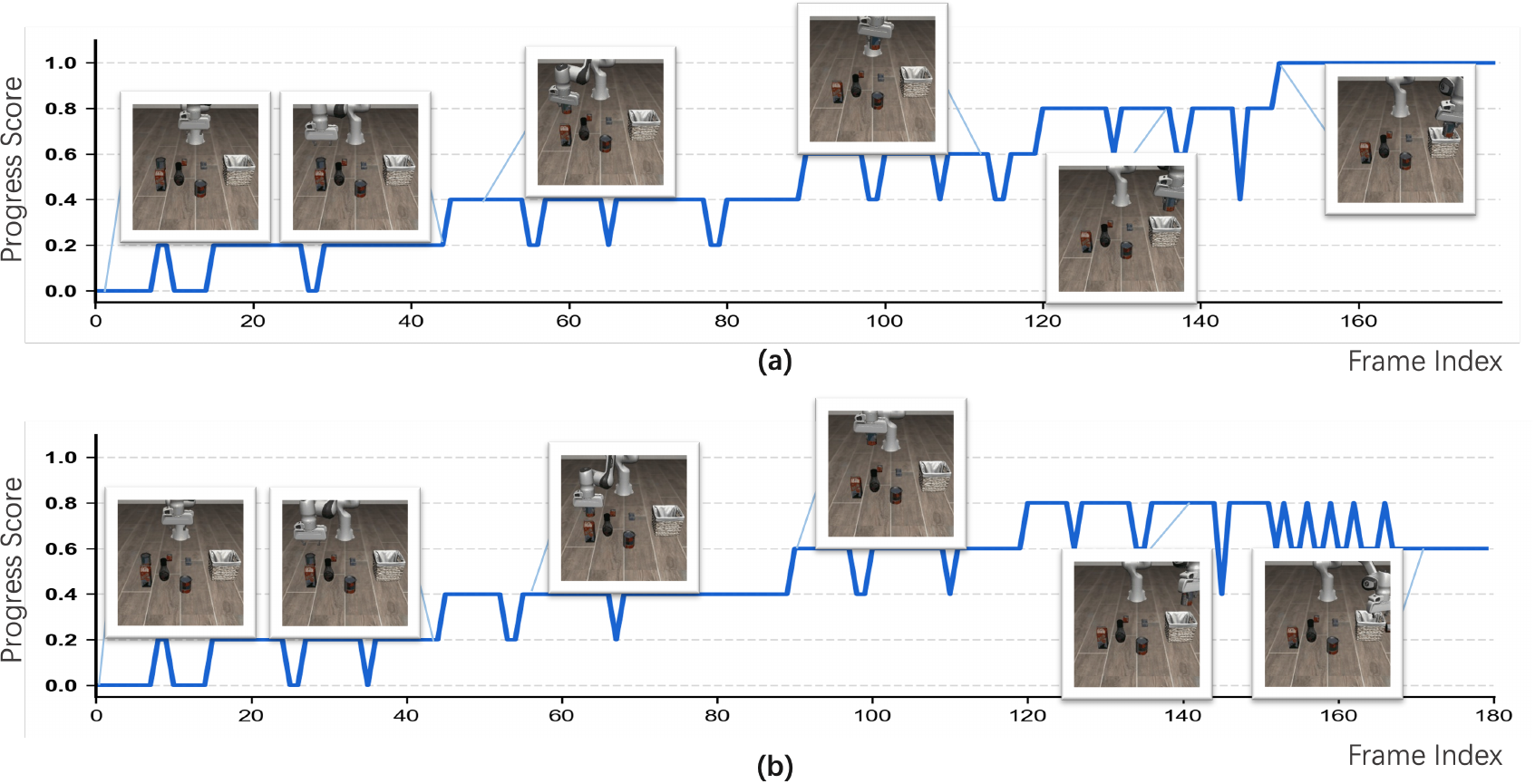}
    \caption{\textbf{Ground-truth Progress Labels.} Visualization of VLM-annotated scores for the LIBERO-Object~\cite{liu2024libero} task \textit{``pick up the alphabet soup and place it in the basket''}. (a) A successful trajectory exhibits step-wise score increases as milestones are achieved. (b) A failure trajectory effectively reflects incomplete execution, with the score stalling at a low value.}
    \label{fig:label_score}
\end{figure*}

\begin{figure*}[t]
    \centering
    \includegraphics[width=0.85\textwidth]{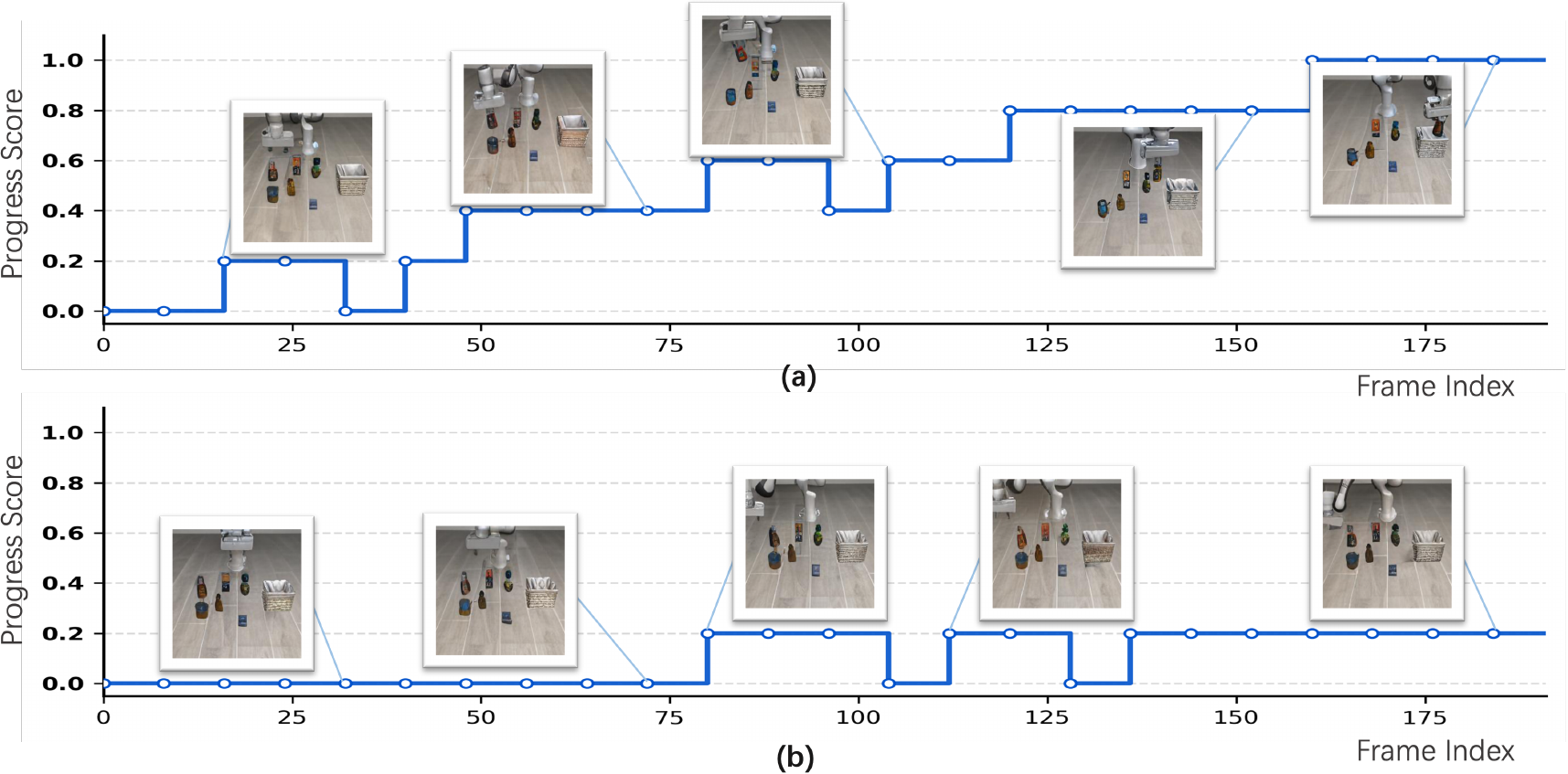}
    \caption{\textbf{Generated Progress Scores.} Visualization of scores predicted by dWorldEval for the LIBERO-Object~\cite{liu2024libero} task \textit{``pick up the ketchup and place it in the basket''}. (a) The model generates a successful rollout where the score accurately rises to 1.0. (b) The model faithfully predicts a failure case, maintaining a low score consistent with the visual outcome.}
    \label{fig:gen_score}
\end{figure*}

\subsection{Visualizing Progress Scores: Labels vs. Generation}
To intuitively demonstrate the effectiveness of our Progress-as-text mechanism, we provide ground-truth annotations and model predictions. 
Figure~\ref{fig:label_score} illustrates the ground-truth labeling process, where the VLM assigns discrete, step-wise scores (e.g., 0.2, 0.4) corresponding to specific achieved milestones. 
Complementing this, Figure~\ref{fig:gen_score} presents the generated progress scores from dWorldEval during inference. 
These results confirm that our model accurately learns to associate visual state transitions with the correct progress semantics.

\section{Visualizing Baseline Consistency}
\label{app:baseline_consistency}

\begin{figure*}[t]
    \centering
    \includegraphics[width=\textwidth]{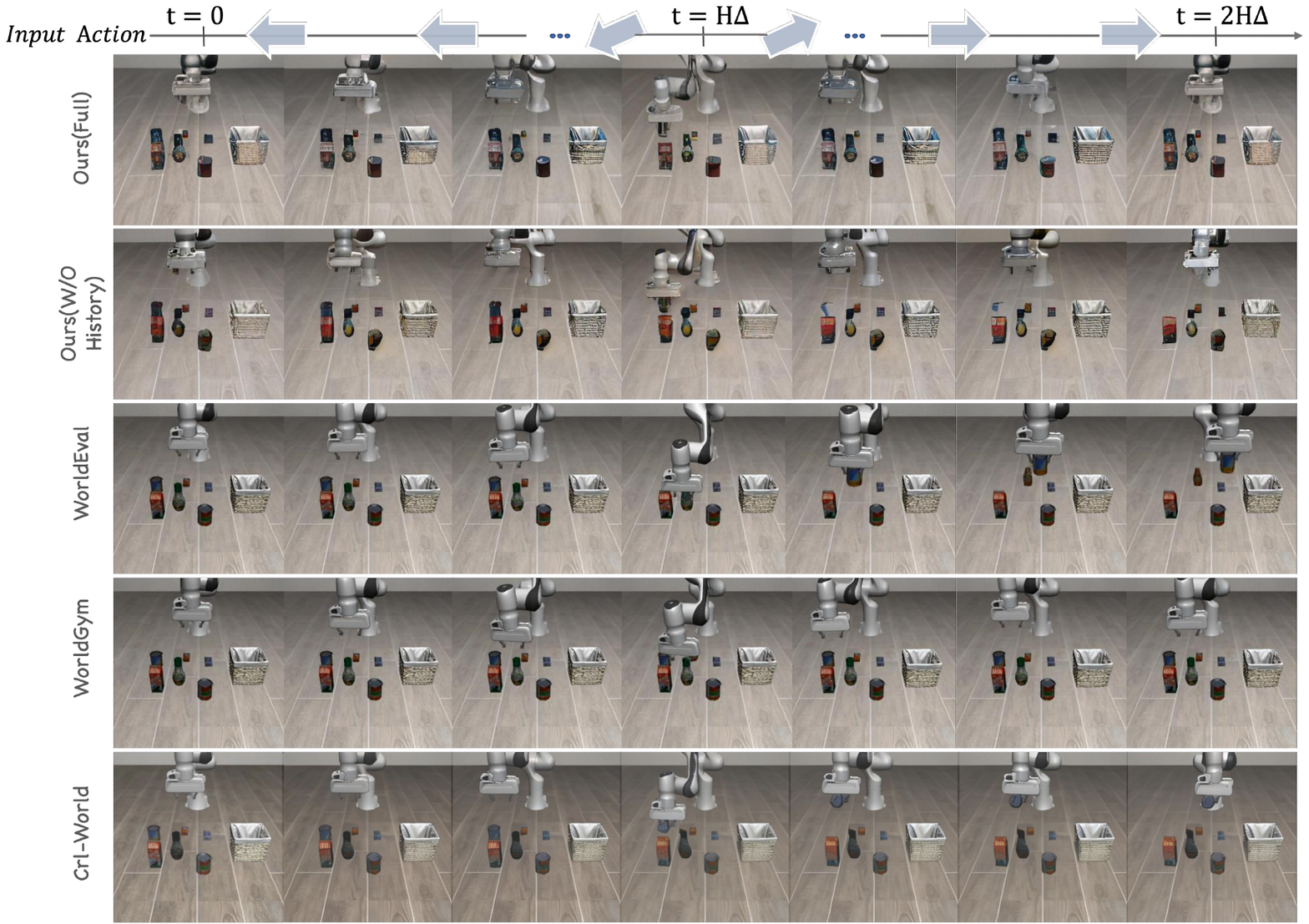}
    \caption{\textbf{Qualitative comparison of round-trip consistency with baselines.} 
    We visualize the round-trip rollouts ($H=20$) on the LIBERO~\cite{liu2024libero} from the shared third-person view. 
    We compare four models: WorldEval~\cite{li2025worldeval}, WorldGym~\cite{quevedo2506worldgym}, Ctrl-World~\cite{guo2025ctrl}, dWorldEval (W/O History), and dWorldEval (Full). 
    The goal is to return to the initial state at $t=2H$ (rightmost column) by executing inverse actions (i.e., reversing the pick-up trajectory to place the object back). Ours (Full) successfully restores the initial scene structure. In contrast, baselines and the w/o-history ablation exhibit significant deviation at the end of the trajectory, caused by a combination of spatiotemporal drift and failure to follow the inverse control signals.}
    \label{fig:baseline_consistency_vis}
\end{figure*}

We provide the qualitative visualization corresponding to the round-trip analysis in Sec.~\ref{sec:more_experiments}.
Figure~\ref{fig:baseline_consistency_vis} illustrates the generated rollouts on the LIBERO~\cite{liu2024libero} suite with a horizon of $H=20$.
It can be observed that dWorldEval successfully restores the initial scene structure at the terminal step $t=2H\Delta$.
In contrast, the baselines (WorldEval and WorldGym) exhibit significant visual deviation from the initial state.
This confirms that the high LPIPS errors reported in the main text stem from a compound failure: the baselines struggle not only with spatiotemporal drift (hallucinating objects) but also with strictly adhering to the inverse action sequence required to return to the start.

\section{More Visualization}
\begin{figure*}[t]
    \centering
    \includegraphics[width=0.9\textwidth]{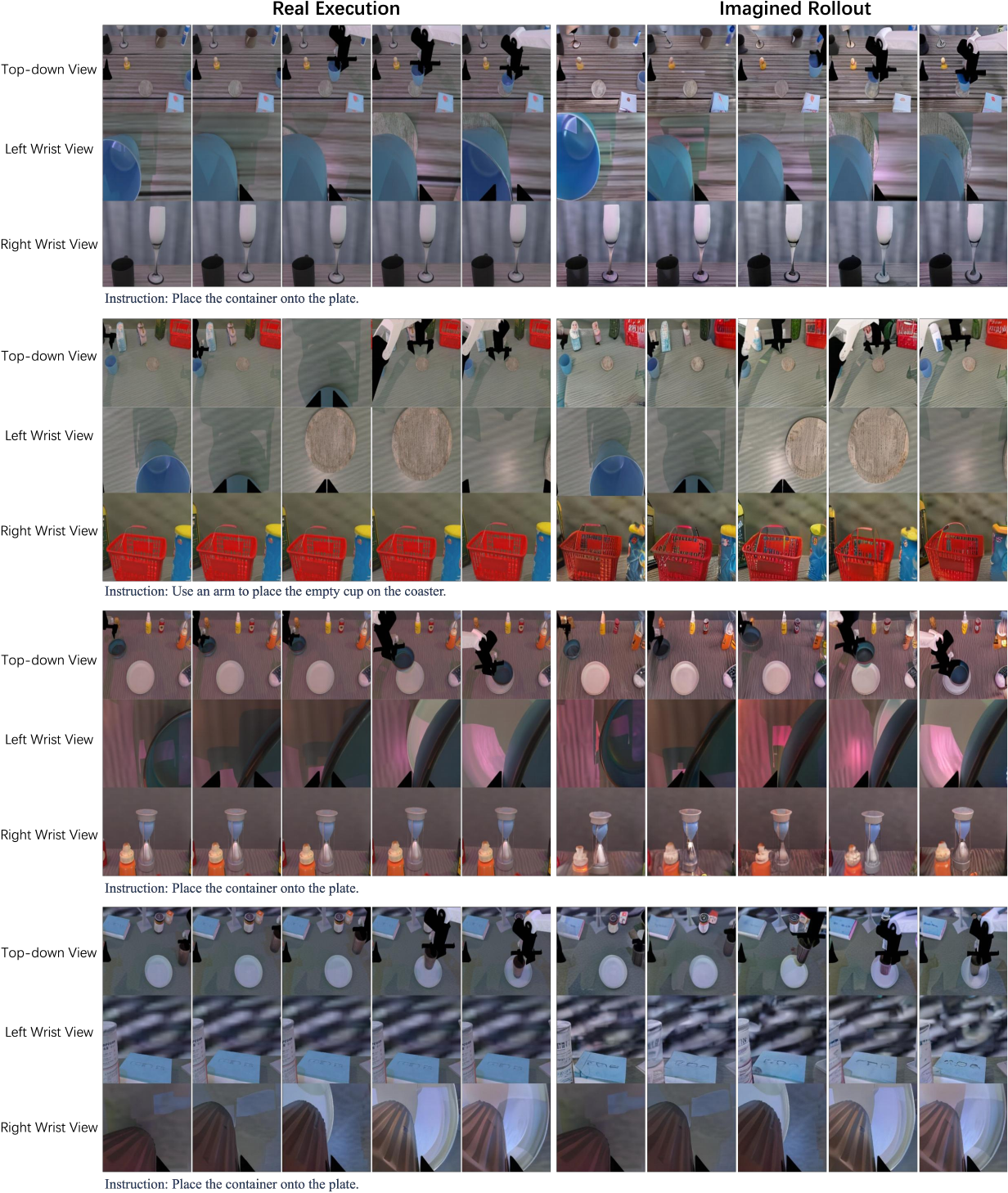}
        \caption{\textbf{Visualization on the RoboTwin benchmark.}~\cite{mu2025robotwin} 
    We compare the ground-truth simulation trajectory with the video generated by our model . 
    Our model synthesizes high-fidelity, synchronized videos across three views (Top-down, Left Wrist, and Right Wrist), accurately preserving the object details and spatial layout of the simulation environment.}
    \label{fig:more_robotwin}
\end{figure*}

\begin{figure*}[t]
    \centering
    \includegraphics[width=\textwidth]{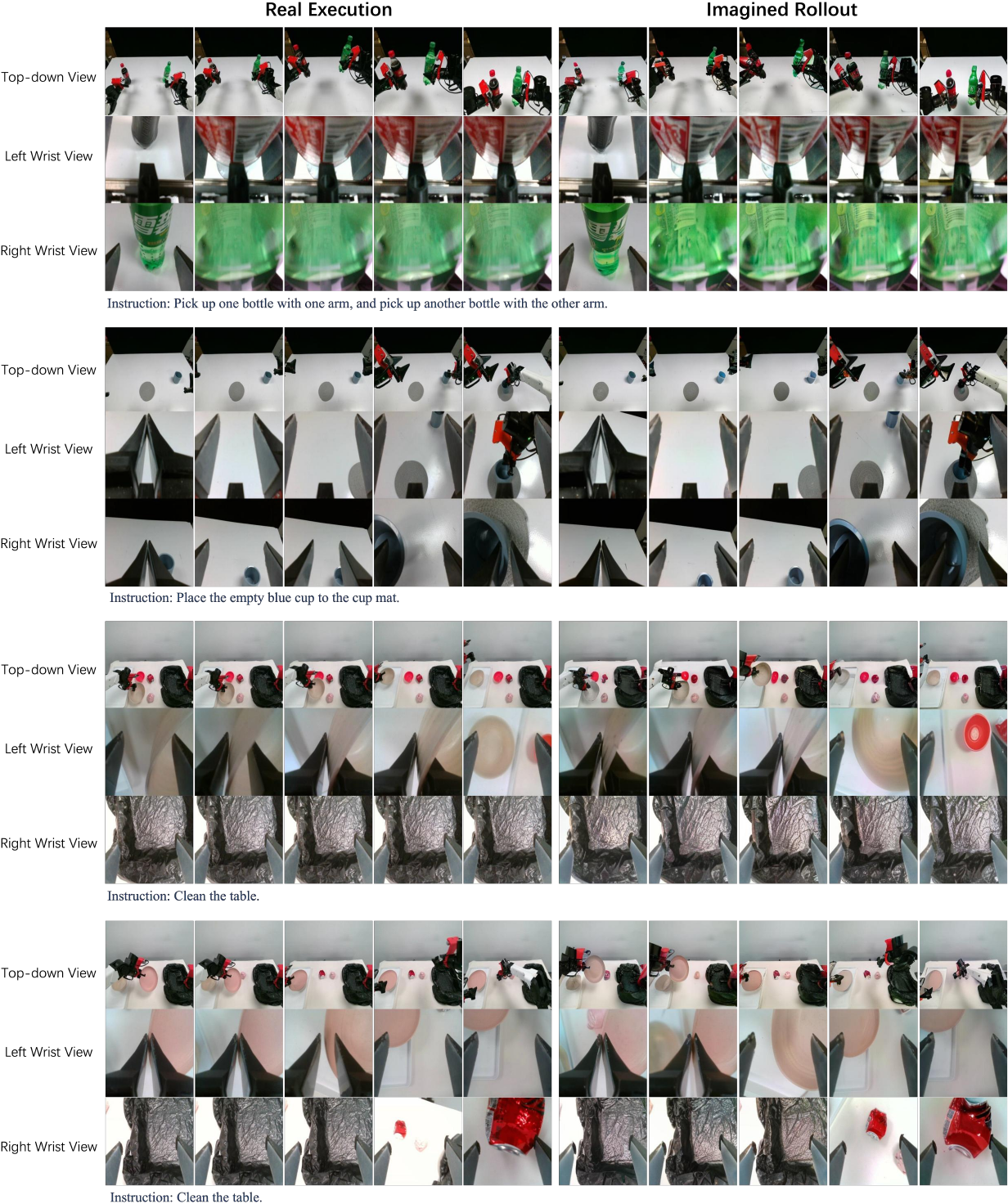}
    \caption{\textbf{Visualization in real-world scenarios.} 
    We compare the ground-truth physical robot execution with the video generated by our model. 
    Given the language instruction, our model synthesizes high-fidelity, synchronized videos across three views (Top-down, Left Wrist, and Right Wrist), accurately preserving object details and handling the visual complexity of the real-world environment.}
    \label{fig:more_real}
\end{figure*}

In this section, we provide more visualizations for both the simulation and the real-world environment. Figure \ref{fig:more_robotwin} presents the generated results on the RoboTwin~\cite{mu2025robotwin} benchmark, and Figure \ref{fig:more_real} shows the real-world cases.

\end{appendices}

\end{document}